\documentclass[runningheads]{llncs}
\usepackage{graphicx}
\usepackage{hyperref}
\usepackage{listings}
\usepackage{amsmath,amsfonts,amssymb}
\usepackage{algorithm}
\usepackage[noend]{algpseudocode}
\usepackage{ulem}
\usepackage[usenames,dvipsnames]{color}
\usepackage[T1]{fontenc}

\begin{document}
\title{Robot at the Mirror: Learning to Imitate via Associating Self-supervised Models}
\titlerunning{Learning to imitate via associating self-supervised models}

\author{
Andrej L\'u\v{c}ny
\and Krist\'ina Malinovsk\'a
\and Igor Farka\v{s}
}
\authorrunning{A. Lúčny et al.}

\institute{
Faculty of Mathematics, Physics and Informatics\\
Comenius University Bratislava, Slovakia\\
\email{\{lucny,malinovska,farkas\}@fmph.uniba.sk}\\
\texttt{http://cogsci.fmph.uniba.sk/cnc/}\\
}
\maketitle
\begin{abstract}
We introduce an approach to building a custom model from ready-made self-supervised models via their associating instead of training and fine-tuning. We demonstrate it with an example of a humanoid robot looking at the mirror and learning to detect the 3D pose of its own body from the image it perceives. 
To build our model, we first obtain features from the visual input and the postures of the robot's body via models prepared before the robot's operation. Then we map their corresponding latent spaces by a sample-efficient robot's self-exploration at the mirror. In this way, the robot builds the solicited 3D pose detector, which quality is immediately perfect on the acquired samples instead of obtaining the quality gradually. The mapping, which employs associating the pairs of feature vectors, is then implemented in the same way as the key--value mechanism of the famous transformer models. Finally, deploying our model for imitation to a simulated robot allows us to study, tune up and systematically evaluate its hyperparameters without the involvement of the human counterpart, advancing our previous research.
\end{abstract}

\section{Introduction}

In nature we observe different forms of skills improvement. Sometimes, it is achieved through gradual learning, for which it is necessary to undergo many repeated attempts \cite{kindsofminds}. At other times we observe that the learning process suddenly occurred based on a single experience. Although we can be amazed by the current achievements of artificial intelligence, the acquisition of most skills is gradual and very lengthy, requiring each behavior pattern to be presented many times. Would it be possible to achieve a sudden improvement in a novel task in just one attempt, given a gradually and slowly prepared set of abilities? This question is especially urgent in mobile robotics, where we already have technical means for running deep learning models on board. However, on-board training or fine-tuning these models is a capacity problem.

In our previous work \cite{itat}, we addressed this issue in an imitation game \cite{imitation} between a human and a humanoid robot iCub \cite{icubsim}. (Please, do not confuse it with the imitation game in the Turing test.) The goal was to teach a robot to imitate a human based on the human imitating the robot. Learning took place in two phases. In phase~1, the robot invited the human to imitate it. The robot created different hand positions, and the human imitated them with his body in front of the robot's camera. It allowed the robot to remember the associations between its body poses and the seen images. In phase~2, the robot imitated a human using the associations acquired in the first phase.

The associations acquired by the robot in phase~1 of the imitation game represent a list of representations of the image the robot sees and the poses the robot has manifested. Technically, ensuring that the robot correctly captures the moment a person takes its pose in the first phase was challenging. However, we simplified this so the person indicated it to the robot by whistling - since his hands are busy taking the right pose while interacting. The second phase of the imitation game relied on the fact that the robot's behavior has a stimulus--response nature. However, it was necessary to solve the problem of using the associations from the first phase because the person will never again be able to take the same pose the robot memorized. Therefore, we needed to design a mechanism to derive the robot's response to a new stimulus from the associations the robot had memorized and, above all, a suitable representation of the image and pose for this mechanism to work.

We use the attention mechanism \cite{attention}, a generally known part of transformers, but in an unconventional way. Our model works with a set of key--value pairs that represent obtained associations. When we have a query at the input, we try to mix it from the available keys and create the output as an analog mixture of the corresponding values.

Associating the image with the pose from the raw data is technically possible, but it does not work. The image data space is too ample, sparse, and fluid to be mapped to another data space by a few associations. Its dimensions are of order hundreds of thousands and contain all possible image inputs, most of which the robot can never see. A slight change in the seen figure can lead to a dramatic shift in the point that represents its picture. Similarly, the pose space of the robot, although much smaller (maximum tens of degrees of freedom), contains many poses that are not reasonable to adopt. A small change in the hierarchically higher degrees of freedom results in a significant difference in the generated posture.

We use deep convolutional networks that process the data into feature vectors to overcome this.
Each feature vector corresponds to a possible image seen or a reasonable pose assumed or, at most, some intermediate form between two such images or poses. Moreover, these spaces are continuous and preserve similarity: the feature vectors corresponding to the gradual change of the seen situation or the adopted posture represent a trajectory in the feature space. Thanks to these properties, we were able to implement the imitation game.\footnote{see the video at \href{https://youtu.be/-3BVbU9BeRE}{https://youtu.be/-3BVbU9BeRE}}

The robot enters the game with two ready-made models: an image encoder and a pose decoder. Both can be obtained without the need for annotation. As an image encoder, we used a pre-trained backbone of a medium-sized self-supervised vision transformer trained by the DINO method \cite{dino}, which encodes the image into 384 features. Next, we obtained the pose decoder by training the variational autoencoder \cite{vae} from a dataset containing the robot's hand movements to randomly selected points around the robot, i.e., obtained during the so-called robot's babbling \cite{imitation2}. Then, in the first phase, the robot generated a pose from several selected posture feature vectors, waited for a signal from the human that it took the correct pose, encoded the image into the features, and saved both feature vectors into the association list. In the second phase, he encoded the image into features, calculated the corresponding pose feature vector from the associations, decoded it, and took the obtained pose. At the same time, the robot could lean towards some memorized posture or combine them appropriately (however, this ability depended on the so-called scaling factor of the association mechanism). An exciting feature of this solution was that the person could deceive the robot in the first phase. The robot learned a lousy reaction if the person did something else instead of the correct pose, for example, showed an object.\footnote{see the video at \href{https://youtu.be/_CBnCOnWRdY}{https://youtu.be/\_CBnCOnWRdY}}

The weakness of this approach was the human involvement, which limited the evaluation of the quality and impact of various system parameters. At the same time, the influence of two parameters was apparent. The first was the number of associations, and the second was the scaling factor of the association mechanism. Another undesirable feature was the need to notify the robot that the associating moment had arrived. 

In this paper, we eliminate these limitations. We train the robot to learn the association between its pose and image in the mirror. It allows us to obtain a (3D) robot pose detector from ready-made self-supervised models without further training or fine-tuning, only based on associating. Then we evaluate it by imitating another robot with the same or similar visage.

Compared to the original solution, we must be able to arrange for the robot to eliminate redundant associations. In addition, our solution solicits both the encoder and decoder for poses. Then, in the first phase, the robot moves in front of the mirror by choosing a random pose feature vector, decoding it, and taking this pose. During the movement of the robot's hands into the new pose, the robot knows, thanks to proprioception, which pose it is currently passing through and can associate each one with the seen image, encoding both into feature vectors and remembering this pair if it is not redundant. The second phase proceeds almost the same way as in the original solution. We will replace the mirror with a view of another robot whose body we can manipulate. Then, by comparing the poses of the two robots, we can evaluate the detector's quality. Unlike within the original imitation game, we can fully automate and assess this process objectively. As a result, we can investigate the parameters' influence and evaluate the limits of the presented approach.

\section{Related work}

Learning by imitation \cite{imitationsurvey} is frequently addressed in cognitive robotics and human--robot interaction. Typically a robot is required to imitate the human companion, as in our previous research mentioned above. 

Research on neural correlates of action understanding, namely the mirror neuron system theory 
suggests that the association between the visual and the motor modality, maintained by these special motor neurons, which also react to visual stimuli, may be the substrate for action understanding or at least mitigate the process of assessing the visual information \cite{tessitore2010}. In our past research, we built a multi-layer connectionist model of action understanding circuitry and mirror neurons, emphasizing the bidirectional activation flow between visual and motor areas in a simulated iCub robot \cite{rebrova2013} and extended the work to perspective-agnostic mirror neurons with results corresponding to biological data \cite{pospichal2019}. The gist of our modeling is to connect the high-level representations of the visual and motor aspects of motor actions in a hetero-associative manner. This allows the robot to understand and replicate the observed action using the motor primitives already in its motor repertoire.

The novel approaches to visual imitation learning usually utilize deep networks that can learn distributions and generate novel samples within, such as the Generative Adversarial Networks (GAN) \cite{gan} and Variational Autoencoders (VAE) \cite{vae}. Generative Adversarial Imitation Learning \cite{liu2} extends the reinforcement learning (RL) paradigm to utilize a smaller expert data sample. Liu et al. utilize GANs and RL to translate the robot’s observation of the demonstration into different contexts, such as different viewpoints, allowing the robot to repeat the observed action. Variational Autoencoders appear even more potent than GANs within this field. Sermanet et al. \cite{sermanet2018} implement imitation learning without any labels utilizing demonstrations in the videos from two different viewpoints yielding a viewpoint-invariant representation of the relationships between the end effectors and the environment with a metric learning loss driving the system to represent the viewpoints for the same action as similar embeddings in the deep model. Similarly, Bahl et al. \cite{bahl2022} propose a system for imitating human actions from the videos recorded in the wild. They base it on reinforcement learning with agent-agnostic representations and conditional VAE employment.

On the path towards imitation via mirroring and building associations between high-level visual and motor modality representations, Zambelli and colleagues \cite{zambelli2020} proposed a multi-modal variational autoencoder to enable the iCub robot to match different modalities up to the point of being able to imitate an observed movement. Further in this line, Seker et al. \cite{seker2022} proposed the new deep modality blending networks (DMBN) with the essence of variational autoencoders, which endows the system to retrieve the missing information of the associated modal information, including different perspectives in the visual data. Garello and colleagues \cite{garello2022} use VAE to map the self-observation and third-person observed perspectives, hence building perspective agnostic representation of actions and using a similar paradigm as our previous MNS research inspired by imitation learning in infants. Namely, the parents tend to involuntarily imitate children right after they produce an action, which could also be a mechanism of the human MNS to emerge as a consequence of Hebbian learning as proposed by Heyes \cite{heyes2010}. 

Šejnová and Štepánová \cite{sejnova2022} utilized the conditional VAE to enable a robot to incrementally learn simple actions from a limited number of demonstrations by a human. Unlike our approach, the labels are presented to the VAE when the robot demonstrates the task. The advantage is that the robot's performance can be assessed during learning. If the particular action receives more training examples, it could start with a minimum of examples and perform a kind of few-shot learning. Marcel and colleagues \cite{marcel2022} use a VAE to model self-touch behavior in developing the body schema in early infancy using a simulated iCub robot with tactile skin. In their work, they iterate through the VAE projections using it as a control loop that will finally produce a movement sequence representing a trajectory from a neutral position to the point of contact of the agent’s arm and its body, just from a single stimulation point. 

Similarly to our current approach, Zahra and colleagues \cite{zahra2022} proposed a two-stage model in which a robot first acquires motor primitives by motor babbling and subsequently learns via imitation. Interestingly, unlike other approaches based on deep learning, they use more biologically relevant spiking networks and self-organizing maps for forming high-level representations of movements similar to our above-mentioned models, which use recurrent self-organizing maps. 

We studied the self-recognition of a robot in the mirror in \cite{mirror}. At that time, we were working with a very simplified representation of the robot body, and it was a big question for us how a robot (or a human) could create a model of its seen body. In this paper, we partially address this question.
 
\section{Our approach}

We aim to make the robot move in front of the mirror and learn to detect its pose from the image it sees. Then we demonstrate the learned ability by imitating the movements of its twin. The twin can be perfect or can vary in textures.

At the same time, we require that learning takes place immediately, based on short-term experience, employing only ready-made models for general image processing and robot poses. Similar to the imitation game mentioned above, we will distinguish two phases. In the first phase, the robot will perceive its image in the mirror, changing due to its babbling movements. In doing so, it gathers sufficient associations between the taken pose and the seen image, but it has to solve the problem with their redundancy. In the second phase, the robot will react to the other robot we can manipulate to take a predefined set of poses. 

\subsection{Ready-made models}

Our solution works with three models: image encoder, pose encoder, and pose decoder. 
The image encoder is the pre-trained backbone of a middle-sized visual transformer trained from a large set of non-annotated images with the DINO method, i.e., in the following self-supervised way. It transforms color images with a resolution of 224$\times$224 into feature vectors of 384 real numbers. Its quality is impressive, demonstrated by several successful applications, including pose detection (of humans). Thus we are almost sure that the vector also contains information representing the robot's pose in the image. But, of course, they are in a very raw form: we use the backbone only, while the applications mentioned above add further processing layers. The model is relatively large, but its middle-sized version can fit into the 4GB GPU. Moreover, its inference only takes 0.05 seconds on an ordinary gaming notebook; thus, it is very suitable for building real-time applications.

We prepare the pose encoder and decoder by training a VAE from a dataset of the proper postures of the robot. We employ the iCub humanoid robot simulator, whose arm contains five significant degrees of freedom, three in the shoulder and two in the elbow joints. Together poses of the left and right arms are coded by ten angles. We collect the dataset using the robot's babbling. We randomly generate points in the robot's vicinity and use inverse kinematics to reach them if possible. Here, inverse kinematics replaces missing feedback that disallows the robot to feel one posture more and another less comfortable. In this way, we have collected 60,000 possible poses of both arms, with the same probability of the robot using the left arm, the right arm, and both arms symmetrically and independently. Then we train the VAE on the dataset. Since the pose space has a low dimension (ten degrees of freedom), we have used just ten input, six intermediate, two feature, six intermediate, and ten output neurons. Of course, the encoder part doubles, generating both the mean and the standard deviation logarithm as typical for VAEs. We have used ReLU and tanh activations since we converted joint angles from $-180^{\circ}$ to $180^{\circ}$ into the range $-1$ to $1$. Before training, we shuffled the dataset and split it into 50,000 training and 10,000 testing examples. The training required ten epochs with batch size 32 and took mere 92 seconds. Finally, we distilled the encoder and decoder parts of the trained model and saved them. Thus the encoder converts ten angles into two features, and the decoder the two features back to the ten angles.\footnote{see the video at \href{https://youtu.be/ZNkF5BTKOLU}{https://youtu.be/ZNkF5BTKOLU}}

\subsection{Association mechanism}

We employ an association mechanism known as attention \cite{attention}. It works with a set of $l$ key--value pairs. When we have a query $q$ as an input, we try to mix it from keys $K$ and create the output as an analog mix from the corresponding values $V$, where
\begin{displaymath}
K = \left({\begin{array}{c} k_1 \\ k_2 \\ \vdots \\ k_l \end{array} }\right)\hspace{20pt}
V = \left({\begin{array}{c} v_1 \\ v_2 \\ \vdots \\ v_l \end{array} }\right)
\end{displaymath}

All queries $q$ and keys $k_l$ are vectors of the dimension $n$, so $K$ is an $l\times n$ matrix. Values $v_l$ and outputs are vectors of the dimension $m$, so $V$ is an $l\times m$ matrix. First, we find $c_i \in \left<0,1\right>$ that $\sum c_i k_i = pr_K(q)$, $\sum c_i =1$, and $i=1,2,...,l$, where $pr_K(q)$ is a vector similar to the projection of $q$ into the subspace generated by the keys $K$. In doing so, we want $c_i$ to express the similarity between the key $k_i$ and the query $q$, so we can derive it from the dot product of $q^T k_i$, proportional to the angle that $q$ and $k_i$ make.

First, however, we have to get these similarities (positive for same, zero for perpendicular, and negative for opposite vectors) to $\left<0,1\right>$, which we can obtain by the function:
\begin{equation}
    {\rm softmax}(x_i)=\frac{\exp(x_i)}{\sum_k \exp(x_k)}
    \label{eq1}
\end{equation}

The coefficients with which we mix the keys $k_i$ into something similar to the query $q$ we, therefore, choose as:
\begin{equation}
    c = {\rm softmax}({\frac{q{K^T}}{d}}), 
    \label{eq2}
\end{equation}    
\noindent where $d$ is a constant that enables us to scale how much we mix from similar keys and how much from different ones. 
The smaller this constant is, the closer the coefficients are to the one-hot encoding. For $d = 1/n$, where $n$ is the dimension of the keys, we always lean towards the dominance of one key, while the value $d = \sqrt{n}$ ensures that we constantly mix a little from the other keys. A proper $d$ can be beneficial for the association mechanism to find the correct response, even for queries for which no similar key was memorized but can be expressed as a transition between two memorized keys. When we have the mixture coefficients $c$, which roughly correspond to the query, we can analogically mix the values of $V$ to the output $o = cV$. So the complete response of the association mechanism $A$ to a query $q$ is calculated as:
\begin{equation} 
 A(q,K,V) = {\rm softmax}\left({\frac{q{K^T}}{d}}\right) V
     \label{eq3}
\end{equation}

The response of the attention mechanism to a query is the same as on its orthogonal projection to the subspace generated by keys:
\begin{equation} 
A(q,K,V) = A(pr^{\rm ort}_K(q),K,V)
     \label{eq4}
\end{equation}
since $q k_l = pr^{\rm ort}_K(q) k_l$. 
This way, the mechanism generalizes when the query does not lie in the subspace generated by keys. Of course, the generalization is as good as the latent space is close to linear.

\subsection{Technical remarks}

For implementation, we need a humanoid robot; we employ iCubSim, the simulator of the iCub robot \cite{icubsim}. We control it from Python via pyicubsim and OpenCV libraries. Further, we have used ONNX runtime for running the image encoder model. We do not need to train it; we have used a pre-trained backbone. We used Keras for training the VAE for postures, dissected its encoder and decoder parts, and converted them from h5 into the pb format for running under OpenCV. We have implemented the association mechanism in NumPy. Since the system operates in real-time, the integration employs a blackboard architecture \cite{agentspace} that helps us to combine slower and faster processes. 

\subsection{Method}

We have a system with an image encoder $F$ (perception), posture decoder $G$ (action), and posture encoder $H$ (proprioception). $F$ transforms input images into feature vectors in the latent space $L_F$, $H$ encodes posture features into $L_{G, H}$, and $G$ decodes them back into the postures. The system has three parameters: the scaling factor $d$ of the association mechanism, the mapping accuracy $\varepsilon$, and a termination condition (number of the collected pairs $t$). We can summarize the system operation into two phases in Algorithm~\ref{alg1}.

{
\begin{algorithm}
\caption{Learning to imitate via association}
\begin{algorithmic}
\State $F$ is image encoder, $G$ is posture decoder, $H$ is posture encoder
\State $A$ is the association mechanism, $K$ keys, $V$ values
\State $d$ is the scaling factor of $A$, $\varepsilon$ is accuracy,
$t$ is the number of collected pairs
\texttt{\\}
\Procedure{phase~1}{$F$,$G$,$H$,$K$,$L$,$d$,$\varepsilon$} \Comment{Learning mirror self-recognition}
\State $K = L = [\hspace{0.1cm}]$ \Comment{start with empty lists of keys and values}
\Loop
\State $i \gets input()$ \Comment{grab the image seen in the mirror}
\State $k \gets F(i)$ \Comment{encode the image into a point in $L_F$}
\State $p \gets proprioception()$ \Comment{get the current posture}
\State $v \gets H(p)$ \Comment{encode the posture into a point in $L_{G,H}$}
\State $w \gets A(k,K,V,d)$ \Comment{potential response $w$ of $A$ to $k$}
\If{$\|v-w\| > \varepsilon$} \Comment{if $w$ differs from $v$ too much}
    \State $K \gets K \cup \{k\}$ \Comment{add $k$ into keys $K$}
    \State $V \gets V \cup \{v\}$ \Comment{add $v$ into values $V$}
    \If{$len(K) = t$} exit \Comment{termination condition}
    \EndIf
\EndIf
\If{${undefined(o)} \vee {p \doteq o}$} \Comment{if the babbling movement is done}
    \State $v \gets random()$ \Comment{generate a point in $L_{G,H}$}
    \State $o \gets G(v)$ \Comment{decode it into angles of the new goal posture}
    \State $output(o)$ \Comment{set the goal posture, i.e., start a new babbling movement}
\EndIf
\EndLoop
\EndProcedure
\texttt{\\}
\Procedure{phase~2}{$F$,$G$,$H$,$K$,$L$,$d$,$\varepsilon$} \Comment{Imitation}
\Loop
\State $i \gets input()$ \Comment{grab the seen image}
\State $q \gets F(i)$ \Comment{encode the image into a point in $L_F$}
\State $v \gets A(q,K,V)$ \Comment{response $v$ of $A$ to $q$}
\State $o \gets G(v)$ \Comment{decode $v$ to the posture $o$}
\State $output(o)$ \Comment{set the posture}
\State 
\EndLoop
\EndProcedure
\end{algorithmic}
\label{alg1}
\end{algorithm}
}

\begin{figure}[ht]
\includegraphics[width=\textwidth]{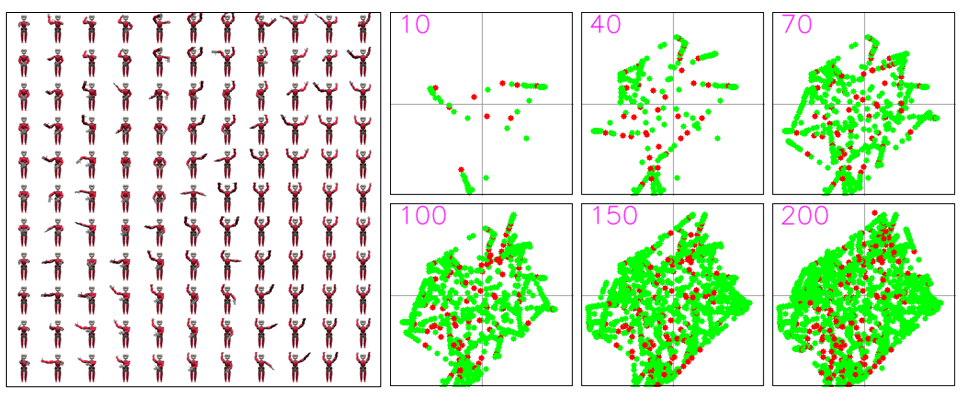}
\vspace{-3mm}
\caption{{\it Left:} Visualization of the pose latent space with topographic organization. {\it Right:} Development of the key--value pairs over time. The red points represent the collected keys, and the green ones are redundant.}
\label{fig1}
\end{figure}

\begin{figure}[ht]
\includegraphics[width=\textwidth]{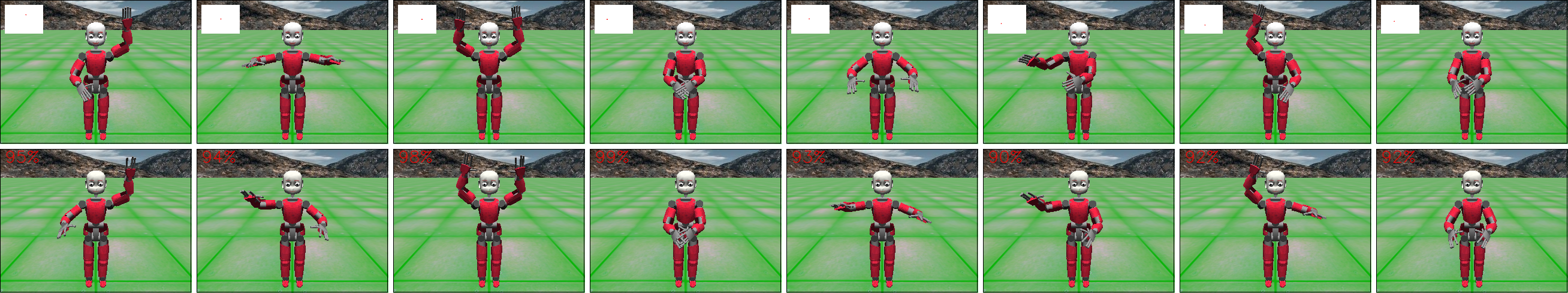}
\vspace{-3mm}
\caption{An example of the learning imitation at the mirror via association. {\it Top:} The testing postures and their points in the latent space. {\it Bottom:} Imitated poses.} 
\label{fig2}
\end{figure}

In phase~1, we start babbling in front of the mirror and gradually collect keys $K$ (image features) and values $V$ (posture features) that provide us with mapping of $L_F$ to $L_{G, H}$. We avoid redundant key--value pairs by checking the response of the association mechanism (Fig.~\ref{fig1}). The babbling aims to reach a random but proper pose as we decode it from random features. In phase~2, we use the collected associations to imitate another robot (Fig.~\ref{fig2}).\footnote{see the video at \href{https://youtu.be/G6xWAKDMpsM}{https://youtu.be/G6xWAKDMpsM}}

\section{Results and Discussion}

Both phases of our algorithm are fully automated so that we can assess its quality objectively (concerning the random nature of the babbling in phase~1). First, we prepare a batch of pose feature vectors corresponding to several good poses (Fig.~\ref{fig2}) that we have not intentionally presented to the robot during phase~1. Then we manipulate another copy of the robot (i.e., we run phase~2), wait until the imitation finishes, and compare the postures of the two robots. Finally, we evaluate the comparison in terms of the normalized mean absolute error (NMAE) calculated as:
\begin{equation}
{\rm NMAE} = {\frac{1}{s} \sum_{j=0}^{s-1} \frac{|{\rm DoF}_j - {\rm DoF}^{\prime}_j|}{{\rm range}_j}}\,\cdot\, 100\%
 \label{eq5}
\end{equation}
where $s=10$ is the number of degrees of freedom, ${\rm DoF}^{\prime}_j$ are joints angles of the imitating robot, ${\rm DoF}_j$ are the angles of its imitated twin, and ${\rm range}_j$ is the angular range of the joint $j$. 
We achieved NMAE of 5.0\% for parameters $d=\sqrt{384}$, $\varepsilon=0.2$, $t=100$. For comparison, if we present exactly eight testing postures to the robot, NMAE decreases to 1.14\%.

\begin{figure}[ht]
\includegraphics[width=\textwidth]{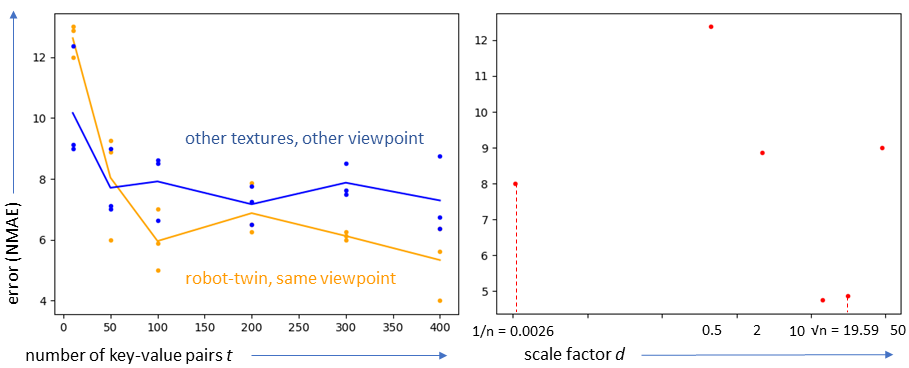}
\caption{ The dependence of NMAE on the number of key-value pairs (left) and on the
scaling factor of the association mechanism (right).} \label{fig3}
\end{figure}

Further, we have investigated the parameters' influence to evaluate this approach's limits. \
First, we tried to modify the number of key--value pairs $t$. A higher number enables us to map the latent spaces more precisely. However, the too-high value decreases the ability of the association mechanism to generalize the mapping. Many irrelevant items are within the $n=384$ features the employed image encoder provides. Therefore $t$ should be significantly lower than the dimension of keys. If they are equal, and the keys are diverse enough, the projection $pr_K(q)$ always equals the query $q$. As a result, there is not much generalization.  For instance, if we change the viewpoint or the robot's color from red to blue, we could fail to recognize its posture. Therefore, the $t$ providing the most stable behavior is about 200 (Fig.~\ref{fig3} left).

Second, we investigated the influence of the scaling factor of the association mechanism $d$. We fixed $t$ and $\varepsilon$ and tried to vary $d$. Lower $d$ like $\frac{1}{n}$ ($n$ is the dimension of keys) achieves a low error for the collected values but approximates the transient postures less accurately. Higher $d$ like $\sqrt{n}$ is less precise for the collected values but generally more suitable (Fig.~\ref{fig3} right).

Finally, we believe the achieved error could be lower if we train a better VAE of poses. During its training, we followed the accuracy given by encoding and decoding the postures. However, it could be profitable also to consider the accuracy of decoding and encoding of the posture feature vectors. 

\section{Conclusion}

In this paper, we investigated our approach to learning imitation by association. We presented an experiment in which a robot learns its posture model from its images seen in the mirror. We designed the procedure such that we could not only test our approach but also be able to objectively evaluate its quality and examine the impact of changing the parameters. 

Our approach is technically interesting, mainly for mobile robots that can use deep learning models on board but lack the capacity for training and fine-tuning. In parallel, from the cognitive science viewpoint, we shed light on body modeling from seen images necessary for performing the imitation task. Namely, we point out that it can emerge quickly, stemming from the gradually developed general models dealing with perception and action separately.

The results of our experimentation provide us with several ideas for further development. We intend to prepare better output models with a more advanced association mechanism in the future. 

We share the code at 
\href{https://github.com/andylucny/learningImitation}{https://github.com/andylucny/learningImitation}

\bigbreak 

\noindent
{\bf Acknowledgements} This work was supported by the EU-funded project TERAIS, no. 101079338, and partly by the national VEGA 1/0373/23 project.

\bibliographystyle{splncs04}
\bibliography{lucny-icann2023}

\end{document}